\documentclass[twoside,11pt]{article}

\usepackage{blindtext}

\usepackage{jmlr2e}

\usepackage{lastpage}
\usepackage{url}            
\usepackage{multirow}
\usepackage{amsmath}
\usepackage{dsfont}
\usepackage{arydshln}
\usepackage{utfsym}
\usepackage{booktabs}       
\usepackage{colortbl}
\usepackage{xcolor}
\usepackage{tikz}
\usepackage{listings} 

\usepackage[framemethod=TikZ]{mdframed}

\definecolor{rowgray}{gray}{0.9} 
\definecolor{mygreen}{HTML}{7baf44}
\definecolor{myred}{HTML}{e22d2d}
\definecolor{myblue}{HTML}{1c88e3}

\definecolor{codegreen}{rgb}{0,0.6,0}
\definecolor{codegray}{rgb}{0.5,0.5,0.5}
\definecolor{codepurple}{rgb}{0.58,0,0.82}
\definecolor{backcolour}{rgb}{0.95,0.95,0.92}

\newmdenv[
    innerleftmargin=10pt,
    backgroundcolor=black, 
    align=center,
    roundcorner=5pt,
]{CodeFrame}

\newcommand{\mythickline}[1]{%
    \tikz[baseline=-0.5ex] \draw[#1, line width=2pt] (0,0) -- (0.3cm,0);%
}

\jmlrheading{23}{2023}{1-\pageref{LastPage}}{12/23}{None}{21-0000}{Shiyu Huang, Wenze Chen, Yiwen Sun, Fuqing Bie and Wei-Wei Tu}

\ShortHeadings{OpenRL: A Unified Reinforcement Learning Framework}{Huang, Chen, Sun, Bie, Tu}
\firstpageno{1}

\begin{document}

\title{OpenRL: A Unified Reinforcement Learning Framework}

\author{\name Shiyu Huang$^{1,\dag}$ \email huangsy1314@163.com
        \AND
       \name Wentse Chen$^2$ \email wentsec@andrew.cmu.edu
       \AND
       \name Yiwen Sun$^3$ \email ywsun22@m.fudan.edu.cn
       \AND
       \name Fuqing Bie$^4$ \email biefuqing@bupt.edu.cn
       \AND
       \name Wei-Wei Tu$^{1,\dag}$ \email tuweiwei@4paradigm.com\\
       \addr $^1$ 4Paradigm Inc., Beijing, China\\
       $^2$ Robotics Institute, Carnegie Mellon University, Pittsburgh, USA\\
       $^3$ School of Data Science, Fudan University, Shanghai, China\\
       $^4$ Beijing University of Posts and Telecommunications, Beijing, China\\
       $^\dag$ corresponding authors
       }

\editor{}

\maketitle

\begin{abstract}
We present OpenRL, an advanced reinforcement learning (RL) framework designed to accommodate a diverse array of tasks, from single-agent challenges to complex multi-agent systems. 
OpenRL's robust support for self-play training empowers agents to develop advanced strategies in competitive settings. 
Notably, OpenRL integrates Natural Language Processing (NLP) with RL, enabling researchers to address a combination of RL training and language-centric tasks effectively. 
Leveraging PyTorch's robust capabilities, OpenRL exemplifies modularity and a user-centric approach. 
It offers a universal interface that simplifies the user experience for beginners while maintaining the flexibility experts require for innovation and algorithm development.
This equilibrium enhances the framework's practicality, adaptability, and scalability, establishing a new standard in RL research. 
To delve into OpenRL's features, we invite researchers and enthusiasts to explore our GitHub repository at \url{https://github.com/OpenRL-Lab/openrl} and access our comprehensive documentation at \url{https://openrl-docs.readthedocs.io}.
\end{abstract}

\begin{keywords}
  Reinforcement Learning, Multi-agent Learning, Self-play, Natural Language Processing, PyTorch
\end{keywords}

\section{Introduction}

Reinforcement learning (RL) has undergone significant expansion in recent years~\citep{arulkumaran2017deep}. This rapidly evolving field has diversified its applications, ranging from single-agent problems~\citep{bellemare2013arcade,lillicrap2015continuous} to the complexities of multi-agent scenarios~\citep{yu2022the, lin2023tizero}, from mastering self-play~\citep{berner2019dota} to the subtleties of offline RL~\citep{fu2020d4rl,qin2022neorl}. These advancements in RL now empower robots to navigate challenging terrains~\citep{polydoros2017survey,margolis2023walk,cheng2023extreme}, enhance large language models in understanding and generating human-like text~\citep{ziegler2019fine,stiennon2020learning,ouyang2022training}, and aid industrial tasks through RL-driven optimization strategies~\cite{roy2021prefixrl,degrave2022magnetic}.
However, this rapid development also presents challenges, as existing RL frameworks struggle to meet the community's growing demands, highlighting the need for an inclusive and adaptable framework.

\begin{table}[t]
\caption{Comparison between OpenRL and other popular RL libraries.}
\begin{center}
        \begin{tabular}{lccccc}
            \toprule
             Library   & NLP & Multi-agent & Self-Play & Offline RL & DeepSpeed\\
            \midrule
            \rowcolor{rowgray}
            {\bf OpenRL}  & \textcolor{mygreen}{$\usym{2713}$} & \textcolor{mygreen}{$\usym{2713}$} & \textcolor{mygreen}{$\usym{2713}$} &  \textcolor{mygreen}{$\usym{2713}$} & \textcolor{mygreen}{$\usym{2713}$} \\
            SB3 &  \textcolor{myred}{$\usym{2717}$} & \textcolor{myred}{$\usym{2717}$} & \textcolor{myred}{$\usym{2717}$} & \textcolor{myred}{$\usym{2717}$} & \textcolor{myred}{$\usym{2717}$} \\
            \rowcolor{rowgray}
            \rowcolor{rowgray}
            Ray/RLlib &  \textcolor{myred}{$\usym{2717}$} & \textcolor{mygreen}{$\usym{2713}$} & \textcolor{mygreen}{$\usym{2713}$} & \textcolor{mygreen}{$\usym{2713}$} & \textcolor{myred}{$\usym{2717}$} \\
            DI-engine &  \textcolor{myred}{$\usym{2717}$} & \textcolor{mygreen}{$\usym{2713}$} & \mythickline{myblue} & \textcolor{mygreen}{$\usym{2713}$} & \textcolor{myred}{$\usym{2717}$} \\
            \rowcolor{rowgray}
            Tianshou &  \textcolor{myred}{$\usym{2717}$} & \mythickline{myblue} & \mythickline{myblue} &  \textcolor{mygreen}{$\usym{2713}$} & \textcolor{myred}{$\usym{2717}$} \\
            MARLlib &  \textcolor{myred}{$\usym{2717}$} & \textcolor{mygreen}{$\usym{2713}$} & \mythickline{myblue} &  \textcolor{myred}{$\usym{2717}$} & \textcolor{myred}{$\usym{2717}$} \\
            \rowcolor{rowgray}
            RL4LMs &  \textcolor{mygreen}{$\usym{2713}$} & \textcolor{myred}{$\usym{2717}$} & \textcolor{myred}{$\usym{2717}$} & \textcolor{myred}{$\usym{2717}$} & \textcolor{myred}{$\usym{2717}$} \\
            trl &  \textcolor{mygreen}{$\usym{2713}$} & \textcolor{myred}{$\usym{2717}$} & \textcolor{myred}{$\usym{2717}$} & \textcolor{myred}{$\usym{2717}$} & \textcolor{mygreen}{$\usym{2713}$} \\
            \rowcolor{rowgray}
            TimeChamber &  \textcolor{myred}{$\usym{2717}$} & \textcolor{myred}{$\usym{2717}$} & \textcolor{myred}{$\usym{2717}$} & \textcolor{mygreen}{$\usym{2713}$} & \textcolor{myred}{$\usym{2717}$} \\
            \bottomrule
        \end{tabular}
    \vspace{-10mm}
    \end{center}
    \label{table:dataset_compare}
\end{table}
In response to this, we introduce OpenRL, a comprehensive framework designed for RL research and development. OpenRL aims to address key challenges in the field: (1) {\bf Comprehensive Integration}: OpenRL embraces a wide spectrum of RL scenarios, including single-agent tasks, multi-agent interactions, and offline RL nuances. Furthermore, OpenRL stands out in two critical emerging areas: the realm of self-play and the convergence of NLP with RL, 
enabling cross-disciplinary research through integrated tools and resources. (2) {\bf Modularity and User-centric Design}: Built on PyTorch~\citep{paszke2019pytorch}, OpenRL features a scalable design with an intuitive interface that melds complexity with ease of use. Our {\bf Gallery} module showcases a set of "reproducibility scripts" for various algorithms and environments. These scripts give novices a clear glimpse into OpenRL's capabilities while serving as a launchpad for experienced users. Complementing these are our detailed documentation and tutorials, ensuring a smooth transition from understanding core concepts to practical implementation.
\begin{figure}[t]
  \centering
  \includegraphics[width=0.85\linewidth]{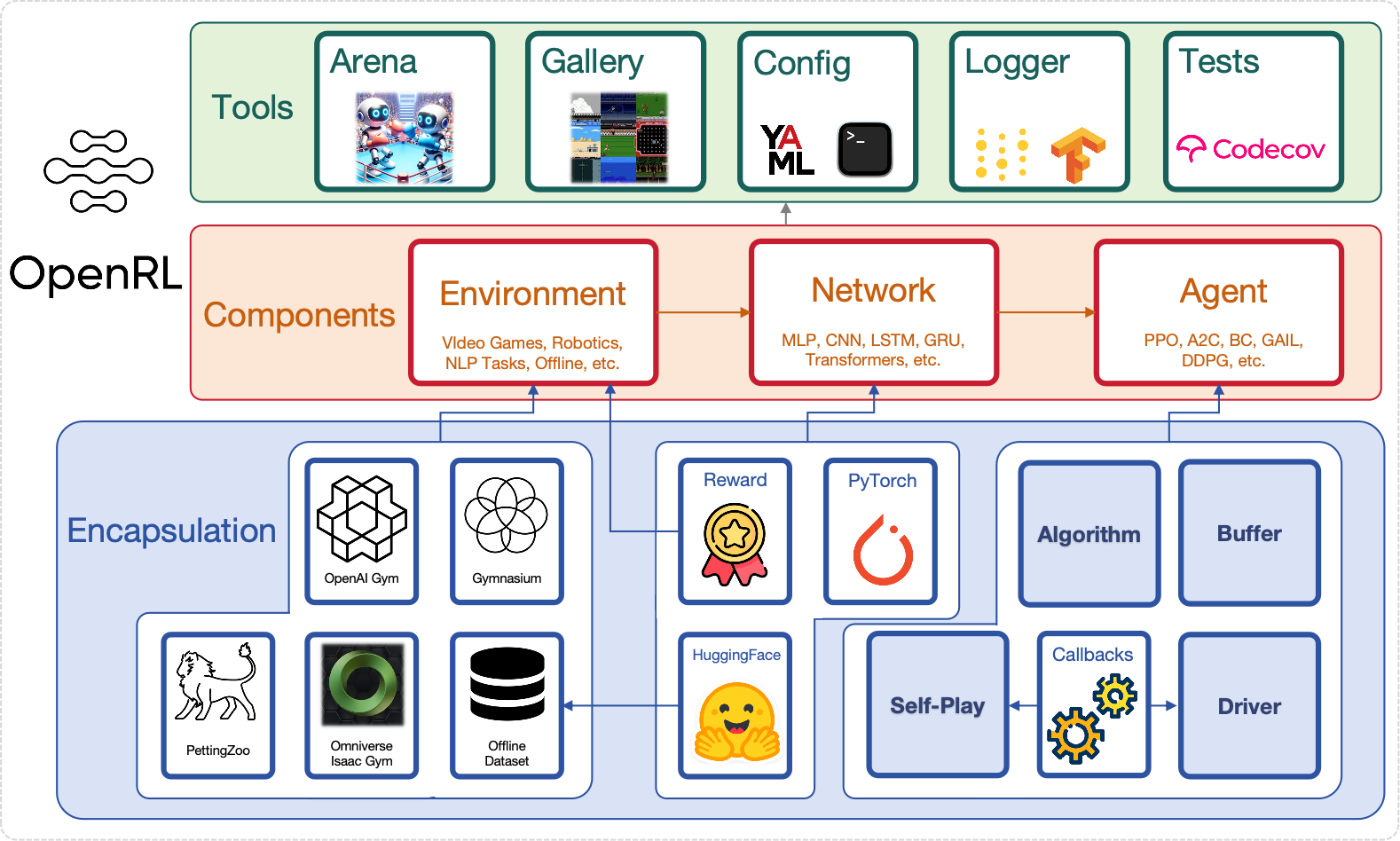}
  \caption{The OpenRL framework is a three-tiered architecture. The \emph{Encapsulation} layer, packages external libraries and essential components required for subsequent modules. The \emph{Component} layer establishes core interfaces for user interaction, including three high-level modules: \emph{Environment}, \emph{Network}, and \emph{Agent}. These modules are instrumental across different tasks and algorithms, sharing a consistent calling interface. The \emph{Tool} layer equips users with multiple utilities to facilitate training, testing, and visualization. OpenRL enables customization of environments, algorithms, and visualizations, offering a versatile and efficient reinforcement learning development framework.}
  \label{fig:framework}
  \vspace{-10mm}
\end{figure}
\section{Design of OpenRL}

OpenRL is an open-source reinforcement learning framework that features a simple, universal interface, supporting training across a wide range of tasks and environments. Figure~\ref{fig:framework} illustrates the overall architecture of OpenRL. The same interface is used for all tasks, including offline reinforcement learning. We present OpenRL's general code interface in Appendix~\ref{ap:interface}. This unified interface simplifies the process of setting up and running reinforcement learning tasks, making OpenRL accessible to both beginners and experienced researchers.

\subsection{Modularity Design}

Modularity is a key aspect of OpenRL's design, contributing significantly to its extensibility. We have abstracted various components of the framework into distinct modules, including the \emph{Reward Module}, \emph{Network Module}, \emph{Algorithm Module}, \emph{Environment Module}, \emph{Agent Module}, and \emph{Buffer Module}. Due to space constraints, we will focus on three main modules: the \emph{Reward Module}, \emph{Network Module}, and \emph{Algorithm Module}.

{\bf Reward Module:} Modifying rewards is a common practice in diverse reinforcement learning algorithms~\citep{eysenbach2018diversity,chen2022dgpo}, exploration algorithms~\citep{burda2018exploration,zhang2020bebold}, and RLHF/RLAIF (RL from AI feedback)~\citep{ouyang2022training,bai2022constitutional}. We provide a base class for the \emph{Reward Module}, which users can extend to easily implement their own reward functions or reward modification functions. 

{\bf Network Module:} We have highly abstracted the network structure, enabling our framework to support various RNN structures such as LSTM~\citep{hochreiter1997long} and GRU~\citep{cho2014learning}, transformers~\citep{vaswani2017attention}, and even networks from other reinforcement learning frameworks. For instance, our framework can load models trained using the Stable Baselines3 framework~\citep{stable-baselines3}.

{\bf Algorithm Module:} We have abstracted the algorithm component to focus on the timing of loss calculation and the combination of losses. In Appendix~\ref{ap:mat}, we provide an example of how to add the MAT algorithm~\citep{wen2022multi} with just a few lines of code.

\subsection{High Performance}

In addition to usability, we also place a strong emphasis on the performance and speed of the OpenRL framework. On a standard laptop, OpenRL can complete the training of the CartPole task in just a few seconds. Compared to the RL4LMs framework~\citep{ramamurthy2022reinforcement}, our training speed for dialogue tasks has improved by $17\%$, with improvements in various performance indicators as well (see Appendix~\ref{ap:nlp} for specific experimental results).

{\bf DeepSpeed:} OpenRL is one of the few reinforcement learning frameworks that support DeepSpeed~\citep{rasley2020deepspeed}, which can help us train larger neural networks in reinforcement learning tasks and accelerate model parallelism through multiple GPUs. We present the training results using DeepSpeed in Appendex~\ref{ap:deepspeed}.

{\bf Mixed Precision Training \& Half-Precision Policy:} We support PyTorch's native mixed precision training to accelerate the training process. During the data collection phase, where randomness is inherently required, the demand for model precision is not as high. Our model supports the use of a half-precision policy for data collection, thereby accelerating the data collection process.

\subsection{Usability}

Designed with usability in mind, OpenRL supports various flexible usage scenarios, including dictionary observation spaces and one-click switching between serial and parallel training environments.

{\bf Configuration:} OpenRL supports setting hyper-parameters through configuration files and command-line parameters. We also support the use of global variables in configuration files, simplifying the process for users to input repeated parameter values.

{\bf Experimental Tracking:} We support the use of Weights \& Biases~\citep{wandb} and TensorBoard~\citep{abadi2016tensorflow} for recording experimental data. We have abstracted this part of the log data, enabling users to record new customized experimental data without modifying OpenRL's underlying code.

{\bf Gallery:} To help users quickly get started with OpenRL and understand its usage, we provide a Gallery module, which includes complete training codes for various algorithms and tasks.

{\bf Arena:} OpenRL provides an Arena module for competitive environments. Agents trained through OpenRL's self-play and rule-based agents can compete and be evaluated in the Arena module.

{\bf Support HuggingFace:} To fully utilize community resources, OpenRL supports loading models and datasets from HuggingFace~\citep{huggingface}. It also supports loading models trained by the Stable Baselines3 organization~\citep{stable-baselines3} on HuggingFace for testing and training.

{\bf Documents:} We provide complete bilingual documentation and detailed tutorials. Our documentation is maintained through the Read the Docs platform, and the documentation address is: \url{https://openrl-docs.readthedocs.io/}.
\section{Conclusion}

In conclusion, OpenRL emerges as a versatile and powerful reinforcement learning framework that offers modularity, high performance, and user-friendliness. Its unique features, such as self-play and NLP integration, make it an indispensable tool for researchers across various levels of expertise. We are dedicated to its continuous enhancement and fostering a collaborative community, confident that OpenRL will drive innovation and cross-disciplinary advances in the dynamic field of reinforcement learning.


\acks{We sincerely thank Fanqi Lin, Ruize Zhang, Ye Li, Geo Jolly and many other open-source contributors for their contributions to the development of OpenRL. We also genuinely appreciate our users for providing numerous constructive comments and suggestions.}


\newpage

\appendix

\section{OpenRL's General Code Interface}
~\label{ap:interface}
The following code snippet provides a general interface for our framework:



\begin{figure}[ht]
  \centering
  \includegraphics[width=0.6\linewidth]{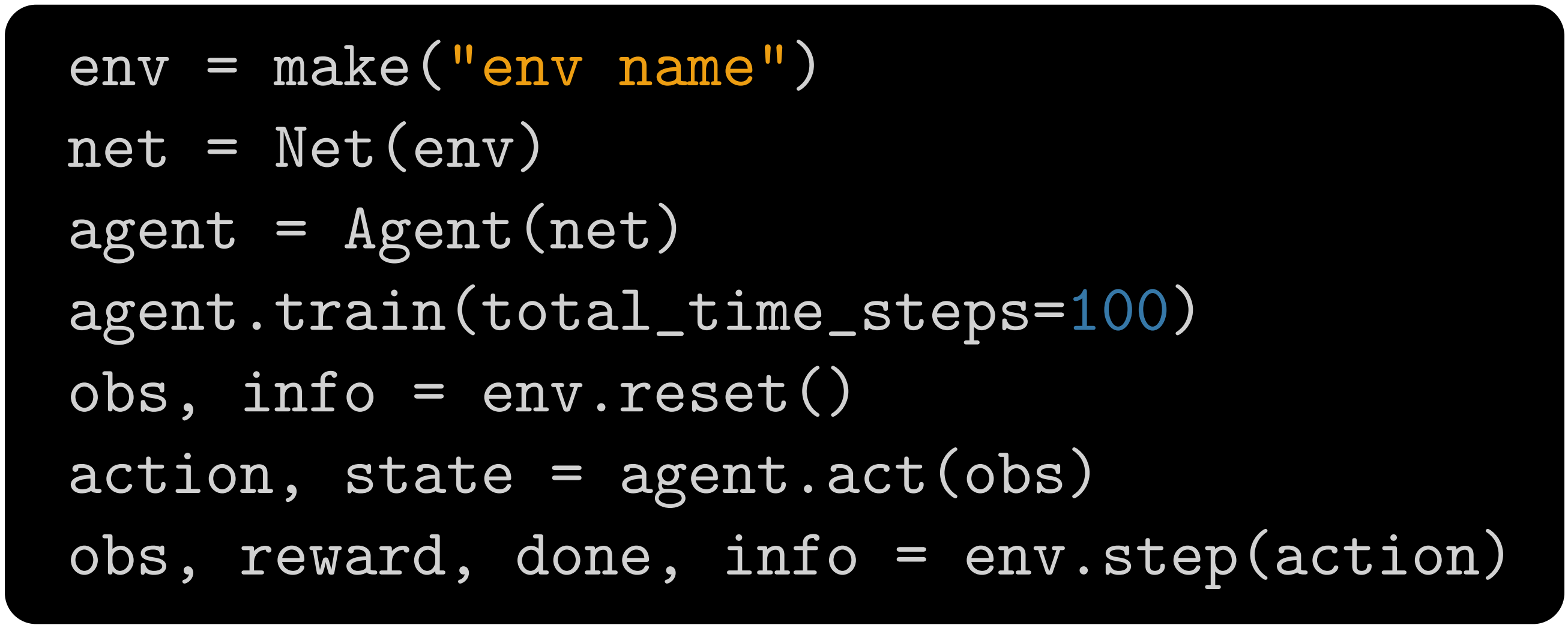}
  \label{fig:code1}
\end{figure}
\section{Implement New Algorithms: Take MAT algorithm as an Example}
\label{ap:mat}

The modular design and abstracted algorithm component of OpenRL simplify the implementation of new reinforcement learning algorithms. Here, we use the MAT algorithm~\citep{wen2022multi} as an example to demonstrate how a new algorithm can be implemented by adding just a few lines of code to the algorithm module:

\begin{figure}[ht]
  \centering
  \includegraphics[width=0.9\linewidth]{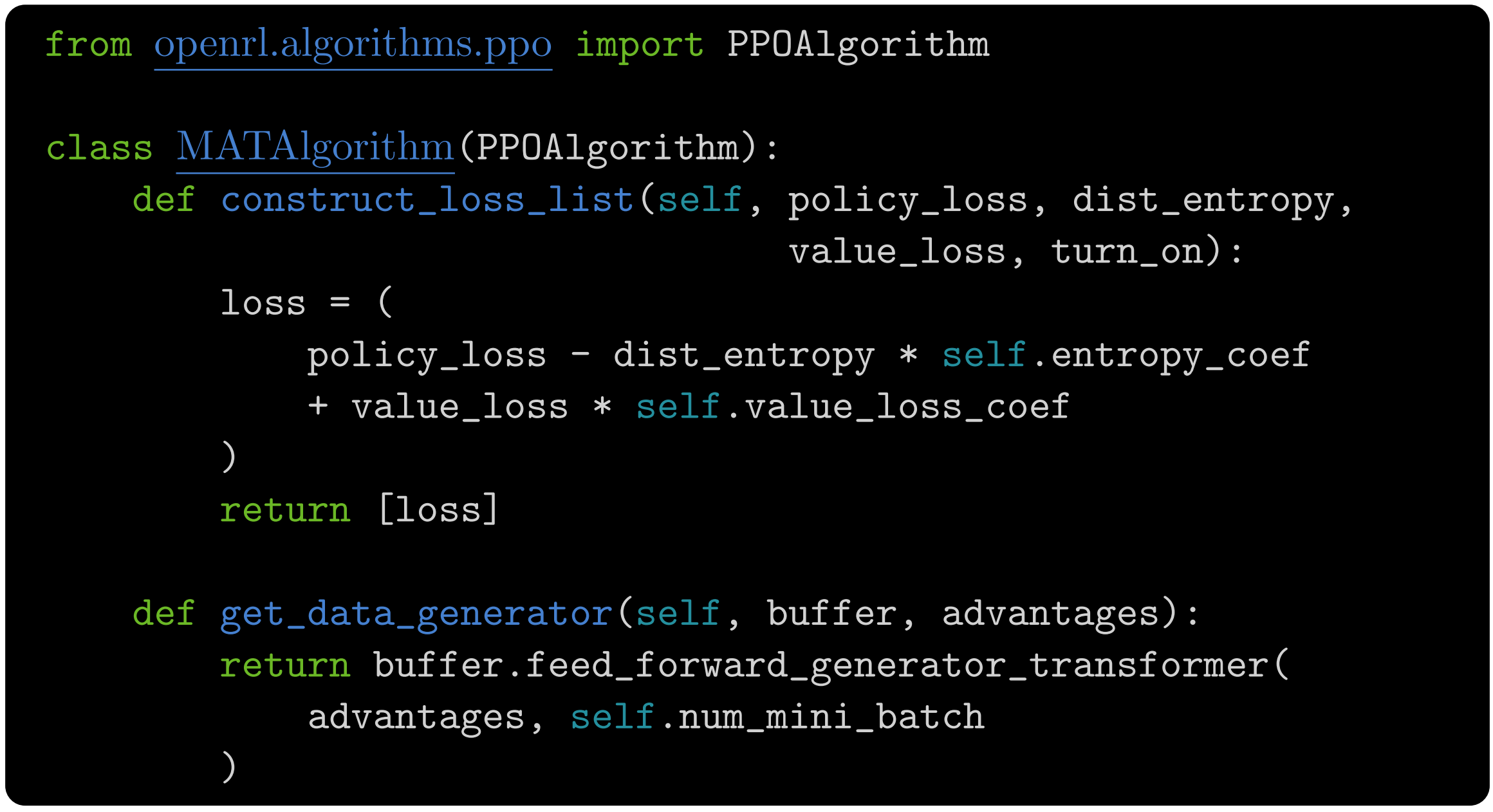}
  \label{fig:code2}
\end{figure}



\section{Training Results Compared to RL4LMs}
\label{ap:nlp}

The table below presents the results of training on the dialogue task~\citep{li2017dailydialog} using OpenRL and comparing them with RL4LMs~\citep{ramamurthy2022reinforcement}. The results indicate that after training with reinforcement learning, all model indicators have improved. Additionally, OpenRL demonstrates a faster training speed compared to RL4LMs (on the same server with NVIDIA 3090 GPUs, the speed is increased by $17.2\%$), as well as better final performance:

\begin{table}[ht]
\centering
\begin{tabular}{lccc}
\hline
\textbf{} & \textbf{OpenRL} & Supervised Learning & RL4LMs \\
\hline
FPS(Speed) & \textbf{13.20(+17\%)} & \textemdash & 11.26 \\
Rouge-1 & \textbf{0.181(+10\%)} & 0.164 & 0.169 \\
Rouge-Lsum & \textbf{0.153(+12\%)} & 0.137 & 0.144 \\
Meteor & \textbf{0.292(+25\%)} & 0.234 & 0.198 \\
SacreBLEU & \textbf{0.090(+43\%)} & 0.063 & 0.071 \\
Intent Reward & 0.435(+1.9\%) & 0.427 & \textbf{0.455} \\
Mean Output Length & 18.69 & 18.95 & 18.83 \\
\hline
\end{tabular}
\caption{Training results on DailyDialog~\citep{li2017dailydialog}.}
\label{tab:dailydialog}
\end{table}

\section{Training Results of DeepSpeed}
\label{ap:deepspeed}

We present in the table below some training results using DeepSpeed~\citep{rasley2020deepspeed} within OpenRL. The outcomes demonstrate that employing OpenRL with DeepSpeed yields faster training speeds compared to using OpenRL with data-parallel.

\begin{table}[h]
\centering
\begin{tabular}{lccc}
\hline
\textbf{} & \textbf{DeepSpeed} & Data-Parallel \\
\hline
FPS(Speed) & \textbf{5.11(+30\%)} & 3.94 \\
Number of GPUs & 2 & 2 \\
Memory Usage per GPU(MB) & 13537 & 7207 \\
GPU Type & RTX 3090 & RTX 3090 \\
Batch Size per GPU & 8 & 8 \\
\hline
\end{tabular}
\caption{DeepSpeed v.s. Data-Parallel for GPT-2-small~\citep{radford2019language}.}
\label{tab:deepspeed_gpt2}
\end{table}

\begin{table}[h]
\centering
\begin{tabular}{lccc}
\hline
\textbf{} & \textbf{DeepSpeed} & Data-Parallel \\
\hline
FPS(Speed) & \textbf{7.09(+35\%)} & 5.25 \\
Number of GPUs & 4 & 4 \\
Memory Usage per GPU(MB) & 35360 & 15854 \\
GPU Type & NVIDIA A100 & NVIDIA A100 \\
Batch Size per GPU & 8 & 8 \\
\hline
\end{tabular}
\caption{DeepSpeed v.s. Data-Parallel for OPT-1.3B~\citep{zhang2022opt}.}
\label{tab:deepspeed_opt}
\end{table}


\bibliography{sample}

\end{document}